\documentclass[sigconf,screen,nonacm]{acmart}

\usepackage{amsmath}
\usepackage{gensymb}
\usepackage{multirow}
\usepackage{subcaption}
\usepackage{fancyvrb}

\AtBeginDocument{%
  \providecommand\BibTeX{{%
    \normalfont B\kern-0.5em{\scshape i\kern-0.25em b}\kern-0.8em\TeX}}}

\begin{document}

%%
%% The "title" command has an optional parameter,
%% allowing the author to define a "short title" to be used in page headers.
\title{Distributed Multi-Agent Deep Reinforcement Learning Framework for Whole-building HVAC Control}

%%
%% The "author" command and its associated commands are used to define
%% the authors and their affiliations.
%% Of note is the shared affiliation of the first two authors, and the
%% "authornote" and "authornotemark" commands
%% used to denote shared contribution to the research.
\author{Vinay Hanumaiah}
\email{vinayha@amazon.com}
\affiliation{%
  \institution{Amazon Web Services}
  \streetaddress{}
  \city{Santa Clara}
  \state{CA}
  \country{USA}
  \postcode{43017-6221}
}

\author{Sahika Genc}
\email{sahika@amazon.com}
\affiliation{%
  \institution{Amazon Web Services}
  \streetaddress{}
  \city{Seattle}
  \state{WA}  
  \country{USA}}

%%
%% By default, the full list of authors will be used in the page
%% headers. Often, this list is too long, and will overlap
%% other information printed in the page headers. This command allows
%% the author to define a more concise list
%% of authors' names for this purpose.
\renewcommand{\shortauthors}{Hanumaiah and Genc, et al.}

%%
%% The abstract is a short summary of the work to be presented in the
%% article.
\begin{abstract}
It is estimated that about 40\%-50\% of total electricity consumption in commercial buildings can be attributed to Heating, Ventilation, and Air Conditioning (HVAC) systems. Minimizing the energy cost while considering the thermal comfort of the occupants is very challenging due to unknown and complex relationships between various HVAC controls and thermal dynamics inside a building. To this end, we present a multi-agent, distributed deep reinforcement learning (DRL) framework based on Energy Plus simulation environment for optimizing HVAC in commercial buildings. This framework learns the complex thermal dynamics in the building and takes advantage of the differential effect of cooling and heating systems in the building to reduce energy costs, while maintaining the thermal comfort of the occupants. With adaptive penalty, the RL algorithm can be prioritized for energy savings or maintaining thermal comfort. Using DRL, we achieve more than 75\% savings in energy consumption. The distributed DRL framework can be scaled to multiple GPUs and CPUs of heterogeneous types.
\end{abstract}

%%
%% The code below is generated by the tool at http://dl.acm.org/ccs.cfm.
%% Please copy and paste the code instead of the example below.
%%
%\begin{CCSXML}
%<ccs2012>
% <concept>
%  <concept_id>10010520.10010553.10010562</concept_id>
%  <concept_desc>Computer systems organization~Embedded systems</concept_desc>
%  <concept_significance>500</concept_significance>
% </concept>
% <concept>
%  <concept_id>10010520.10010575.10010755</concept_id>
%  <concept_desc>Computer systems organization~Redundancy</concept_desc>
%  <concept_significance>300</concept_significance>
% </concept>
% <concept>
%  <concept_id>10010520.10010553.10010554</concept_id>
%  <concept_desc>Computer systems organization~Robotics</concept_desc>
%  <concept_significance>100</concept_significance>
% </concept>
% <concept>
%  <concept_id>10003033.10003083.10003095</concept_id>
%  <concept_desc>Networks~Network reliability</concept_desc>
%  <concept_significance>100</concept_significance>
% </concept>
%</ccs2012>
%\end{CCSXML}
%
%\ccsdesc[500]{Computer systems organization~Embedded systems}
%\ccsdesc[300]{Computer systems organization~Redundancy}
%\ccsdesc{Computer systems organization~Robotics}
%\ccsdesc[100]{Networks~Network reliability}

%%
%% Keywords. The author(s) should pick words that accurately describe
%% the work being presented. Separate the keywords with commas.
\keywords{Reinforcement learning, deep learning, HVAC optimization, machine learning}

%%
%% This command processes the author and affiliation and title
%% information and builds the first part of the formatted document.
\maketitle

%%
%% The acknowledgments section is defined using the "acks" environment
%% (and NOT an unnumbered section). This ensures the proper
%% identification of the section in the article metadata, and the
%% consistent spelling of the heading.
%\begin{acks}
%To Robert, for the bagels and explaining CMYK and color spaces.
%\end{acks}

\section{Introduction}

Buildings account for 40\% of total energy consumption, 70\% of total electricity, and 30\% of carbon emissions in the United States~\cite{shaikh:rser14, doe:energy11}. HVAC systems account for 50\% of the total energy consumption in buildings. The aim of HVAC systems in residential and commercial buildings is to maintain indoor air temperature and air quality. The conventional building HVAC control is rule-based feedback. In this setup, temperature setpoints are set based on certain pre-determined schedules (e.g. day and night time schedules). A simple controller like Proportional-Integral-Derivative (PID)~\cite{geng:icca93} is used for tracking the setpoints. This simple reactive strategy works well to maintain indoor air temperature on a per-zone basis, but would not be optimal for a large building with many thermal zones. The strategy also ignores the effects of external elements like weather. Additional savings in energy can be achieved by intelligently controlling HVAC systems.

Optimal control strategies, such as MPC (Model Predictive Controller) address above limitations by iteratively optimizing an objective function over a finite time horizon. One of the limitations of MPC is the need for accurate models of the environment. The thermal dynamics in a large building is very complex with zones at the periphery having different thermal properties than one in the middle, and similar behavior can be seen with multiple floors in the building. Moreover, the dynamics of two buildings can be very different depending on the layout, HVAC configurations, and occupancy rates. Modeling these different thermal properties of buildings is difficult and needs to be repeated for a new building. Model-free optimization is preferred in such scenarios and Reinforcement learning (RL)~\cite{sutton:rl18} is well suited for model-free optimization. RL directly interacts with an environment, and learns the model through set of actions and corresponding feedback in the form of state changes. DRL incorporates deep learning with RL, allowing agents to make decisions from unstructured input data without manual engineering of the state space. DRL has achieved remarkable success in playing Atari and Go~\cite{mnih:nature15}. DRL is especially well suited for model-free RL, where the agent can learn to model the environment by exploring extensively. Ray RLlib~\cite{liang:arxiv18} is a popular DRL framework, which supports commonly used DRL algorithms.

Since RL algorithms require extensive action-state pairs from an environment to optimize, RL algorithms are usually trained on simulators, as (i) trying those actions in real world could be dangerous, e.g. setting temperature too high or cold in a building, and (ii) collecting the extensive data would take a long time in real world. EnergyPlus~\cite{crawley:energplus01} is a whole building energy simulation program that models heating, cooling, ventilation, lighting, and water use in buildings. EnergyPlus is commonly used by design engineers and architects, who would like to design for HVAC equipment, or analyzing life cycle cost, or optimize energy performance, etc. 

In this paper, we present a co-simulation framework, where DRL algorithms from Ray RLlib will interact with EnergyPlus simulator to arrive at optimal HVAC policy. The co-simulation framework will provide seemless and scalable interface, where observations from EnergyPlus are provided to OpenAI gym environment, which in turn will compute reward. The rewards are collected by RLlib and compute the optimal actions for next time step. These actions are conveyed back to OpenAI gym and is supplied to EnergyPlus. Though this co-simulation framework, EnerygPlus, which is designed to run as single process, is extended RLlib's distributed model to scale for multiple CPUs and GPUs across clusters. This is made possible by using callbacks from EnergyPlus and utilizing queues to synchronize the two softwares. Additionally, we demonstrate setting up multi-zone and multi-agent environments with this framework.

The main contributions of this work are summarized below:
\begin{enumerate}
	\item This paper provides an optimization framework by orchestrating a co-simulation environment between Ray RLlib and EnergyPlus. The framework is highly scalable to using multiple compute instances including CPUs and GPUs.
	\item The above framework uses standard OpenAI gym, which is customizable, and supports multiple DRL algorithms (all that is supported by RLlib). We demonstrate setting up multi-zone control and multi-agents to control building at different times of the year.
	\item We conduct experiments for different weather conditions and simulator configurations. We demonstrate the trade-off in training times and rewards w.r.t. energy-temperature penalty coefficient.
\end{enumerate}

\subsection{Related Work}

The recent literature in HVAC optimization for buildings generally falls in two categories. One that uses MPC, the other uses RL. One of the roadblocks in widespread adoption of MPC is the need for a model~\cite{privara:cca11, killian:be16}. Due to heterogeneity of the buildings, we may need to develop model for each thermal zone~\cite{lu:appEnergy15}. The models could be built based on physics, e.g. EnergyPlus, or based on material characteristics, e.g. computation fluid dynamics models. These models are not control-oriented~\cite{ercan:icbs16}, and although not impossible, it requires considerable work to control these models.

Many DRL based HVAC control methods have been proposed. Wei et al.~\cite{wei:dac17} DRL method based on Deep Q-Network (DQN)~\cite{mnih:arxiv13} was one of the first DRL based control methods to minimize energy consumption, while maintaining temperature in a desired range. Gao et al.~\cite{gao:arxiv19} proposed a Deep Deterministic Policy Gradients (DDPG)~\cite{lillicrap:arxiv19} based DRL method to minimize energy consumption and thermal discomfort in a laboratory setting. Similarly, Zhang et al.~\cite{zhang:eb19} proposed Asynchronous Advantage Actor Critic (A3C)~\cite{mnih:arxiv16} based control to jointly optimizes energy demand and thermal comfort in an office building. 

Some of the works have utilized EnergyPlus to model the building and use it as a simulator to verify the HVAC control methods. Chen et al.~\cite{chen:arxiv19} validated their work on a large office (16 thermal zones) within EnergyPlus by controlling the temperature setpoints. Moriyama et al.~\cite{moriyama:asc18} presented a test bed that integrates DRL with EnergyPlus for data center HVAC control. Our work is most similar to this work. One of the limitation of~\cite{moriyama:asc18} was the co-simulation is not scalable to beyond one core. In this paper, we provide the framework that is highly scalable to multiple cores and multiple nodes. 

The rest of the paper is structured as follows: Section~\ref{sec:model} covers problem formulation and introduction to EnergyPlus simulator. Section~\ref{sec:drl} introduces the reader to deep reinforcement learning algorithms for HVAC control. Section~\ref{sec:experiments} presents the results of exploring DRL algorithms with EnergyPlus, and finally we conclude the paper in Section~\ref{sec:conclusion}.
\section{HVAC System Modeling of Buildings}
\label{sec:model}

In this section, we will cover the problem formulation, discuss EnergyPlus and reinforcement learning frameworks, and strategies to distribute workloads to speed up computation.

\subsection{Problem Formulation}

Table~\ref{tab:nomenclature} lists the nomenclature used in this paper:

\begin{table}[ht]
	\caption{Nomenclature used in the paper.}
	\label{tab:nomenclature}
	\begin{tabular}{lp{2.65in}}
		\toprule
		Symbol & Description\\
		\midrule		
		$N_z$ & Number of thermal zones.\\
		$N_d$ & Number of simulation days.\\
		$t_s$ & Timestep (s)\\
		$T_{min}, T_{max}$ & Minimum and maximum acceptable zone temperature ($\degree$ C), respectively\\
		$T_z$ & Zone $z$ temperature ($\degree$ C), $\mathbf{T}$ represents the vector of all zone temperatures.\\
		$P_z^h$ & Heat added to zone $z$ per unit timestep\\
		$P_z^c$ & Heat removed from zone $z$ per unit timestep\\		
		$P_b$ & Base energy expenditure (unrelated to zone) per unit timestep, e.g. energy used for lighting, etc.\\
		$\alpha$ & Energy-temperature penalty ratio\\
		$T_z^h$ & Heating setpoint temperature for zone $z$ ($\degree$ C)\\
		$T_z^c$ & Cooling setpoint temperature for zone $z$ ($\degree$ C)\\	
		$T_e$ & Outdoor temperature ($\degree$ C)\\	
		$R_h$ & Relative humidity (\%)\\			
		$T^h_{min}, T^h_{max}$ & Minimum and maximum heating setpoint ($\degree$ C), respectively\\
		$T^c_{min}, T^c_{max}$ & Minimum and maximum cooling setpoint ($\degree$ C), respectively\\						
		\bottomrule
	\end{tabular}
\end{table}

The goal of our work is to minimize energy consumption in a building subject to maintaining zone temperatures within the range of comfort. The problem can be mathematically stated as follows:
\begin{align}
	\displaystyle\min_{T_z^h(t_s),~T_z^c(t_s)} \quad & \displaystyle\sum_{t_s=0}^{t_s=N_d} \left[ P_b(t_s) +  \displaystyle\sum_{z=0}^{z=N_z} P_z^h(T_z^h, t_s) + P_z^c(T_z^c, t_s) \right]  \label{eqn:orig-obj}\\
	s.t. \quad & T^h_{min} \le T_z^h(t_s) \le T^h_{max}.  \quad \forall z, t_s \label{eqn:t_h_range}\\
	& T^c_{min} \le T_z^c(t_s) \le T^c_{max}. \quad \forall z, t_s \label{eqn:t_c_range}\\		
	& T_z(t_s) = f_T(\mathbf{T}, T_e, R_h, \mathbf{P}^h, \mathbf{P}^c, t_s) \quad \forall z, t_s \label{eqn:t_func}\\
	& P_z^h(t_s) = f_{Ph}(\mathbf{T}, T_z^h, t_s), \quad \forall z, t_s \label{eqn:p_h_func}\\	
	& P_z^c(t_s) = f_{Pc}(\mathbf{T}, T_z^c, t_s), \quad \forall z, t_s \label{eqn:p_c_func}\\		
	& T_{min} \le T_z(t_s) \le T_{max}. \quad \forall z, t_s \label{eqn:t_range}
\end{align}

Equation~\ref{eqn:orig-obj} is the objective function, where we minimize the overall energy consumption for all simulation steps and all zones. Power consumed for heating and cooling are described in Equations~\ref{eqn:p_h_func} and~\ref{eqn:p_c_func}, respectively. The power consumed is a function of the setpoint, the temperature of all zones. Since temperature of a zone is influenced by other nearby zones, we need temperature of all zones to compute the effective power to bring the temperature of a zone to it's desired range. Similarly, function for computing zone temperatures is dependent on both heating and cooling setpoints, and the temperature of all zones, as described in Equation~\ref{eqn:t_func}. Note that $ f_{Ph}$, $ f_{Pc}$, and $ f_T$ are black-box functions, which are learned by Model-free RL algorithms described in Section~\ref{sec:drl} as they optimize for reward.

Of the constraints mentioned in Equations~\ref{eqn:t_h_range},~\ref{eqn:t_c_range}, and~\ref{eqn:t_range}, Equation~\ref{eqn:t_range} is the constraint on the observed temperatures, while the rest are the constraints on the input variables. Equation~\ref{eqn:t_range} presents a challenge in optimizing for algorithms, as solving for hard constraint increases the complexity of the algorithms. Equations~\ref{eqn:t_h_range} and~\ref{eqn:t_c_range} are constraints on control variables, which are easily handled by limiting the range of input variables. One common way of addressing this issue is by modifying the hard constraint as a soft constraint in the objective with an added penalty coefficient as shown in Equation~\ref{eqn:obj}.
\begin{multline}
	\displaystyle\min_{T_z^h(t_s),~T_z^c(t_s)} \quad \displaystyle\sum_{t_s=0}^{t_s=N_d} \left[ P_b(t_s) +  \displaystyle\sum_{z=0}^{z=N_z} \left( P_z^h(T_z^h, t_s) + P_z^c(T_z^c, t_s) \right. \right. \\
	+ \alpha (\max(T_{min} - T_z(t_s), T_z(t_s) - T_{max}, 0))^\lambda)]  \label{eqn:obj}
\end{multline}	
\begin{align}
	s.t. \quad & T^h_{min} \le T_z^h(t_s) \le T^h_{max}.  \quad \forall z, t_s\\
	& T^c_{min} \le T_z^c(t_s) \le T^c_{max}. \quad \forall z, t_s\\		
	& T_z(t_s) = f_T(\mathbf{T}, T_e, R_h, \mathbf{P}^h, \mathbf{P}^c, t_s) \quad \forall z, t_s\\
	& P_z^h(t_s) = f_{Ph}(\mathbf{T}, T_z^h, t_s), \quad \forall z, t_s\\	
	& P_z^c(t_s) = f_{Pc}(\mathbf{T}, T_z^c, t_s), \quad \forall z, t_s\\		
	& \alpha, \lambda \ge 0, \label{eqn:alpha_const}\\
	& \lambda \ge 1 \label{eqn:lambda_const}
\end{align}

We have added a second term in Equation~\ref{eqn:obj} to represent the violation of zone temperature constraints. $\max(T_{min} - T_z(t_s), T_z(t_s) - T_{max}, 0)$ is the max of violation of zone temperature of either below minimum and above maximum temperature. We also add two coefficients, $\alpha$, which prioritizes reducing power consumption vs temperature comfort. Increasing $\alpha$ penalizes more on violating temperature constraints, and decreasing $\alpha$ will give more emphasis on energy reduction. $\alpha$ is constrained to be above $0$ as in Equation~\ref{eqn:alpha_const}, but with no upper limit. The reason being that the power consumption and temperature are of different scales and cannot be normalized to same range.

\begin{figure}[ht]
	\centering
	\includegraphics[width=0.9\linewidth]{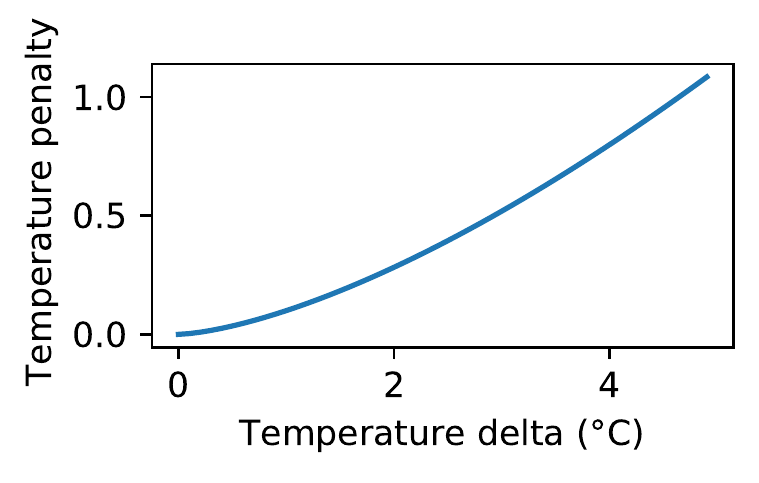}
	\caption{Temperature penalty curve adopted in our experiments.}
	\label{fig:temp-penalty}	
	\Description{RL: Training cost}
\end{figure}

$\lambda$ is a pre-determined constant that determines the penalty as a function of deviation of zone temperature from either minimum or maximum constraint. Increasing $\lambda$ is exponentially increases penalty. It's constrained to be more than $1$ as in Equation~\ref{eqn:lambda_const}. An example of temperature penalty term for $\alpha=1$ and $\lambda=1.5$ is shown in Figure~\ref{fig:temp-penalty}. Temperature delta refers to the delta by which zone temperature exceeding the temperature constraints.

In Equation~\ref{eqn:obj}, we added a new penalty coefficient $\alpha$, which prioritizes reducing power consumption vs temperature comfort. Increasing $\alpha$ penalizes more on violating temperature constraints, and decreasing $\alpha$ will give more emphasis on energy reduction. $\alpha$ is constrained to be above $0$ as in Equation~\ref{eqn:alpha_const}, but with no upper limit. The reason being that the power consumption and temperature are of different scales and cannot be normalized to same range. Rest of the equations remain same as Equations~\ref{eqn:t_h_range} --~\ref{eqn:p_c_func}.

The following sections discuss the simulation engine we use for our experiment, and the co-simulation with DRL optimizers. 

\subsection{Ray RLlib and OpenAI Gym}

RLlib~\cite{liang:arxiv18} is a popular open-source library for reinforcement learning (RL) that is built on Ray~\cite{moritz:arxiv18}. RLlib offers highly scalability and packs many commonly used policies, e.g. PPO~\cite{schulman:arxiv17-ppo}, DDPG~\cite{lillicrap:arxiv19}, etc., for RL training. Gym~\cite{brockman:corr16} is a toolkit from OpenAI for developing reinforcement learning algorithms. Gym provides standard interfaces for initializing environment, where we define the action and observation spaces, functions for resetting environment, and providing controls and observing outputs at every time step of the simulation. RLlib interfaces with Gym seamlessly. 

\subsection{Co-simulation Routine}

We combine EnergyPlus with a RL framework to co-simulate and determine the optimal settings for HVAC operation. A user can provide building information, e.g. HVAC system (heat sources, rate of cooling/heating, etc.), number of zones, building material, etc., in an input file called IDF. EnergyPlus reads the IDF file and the weather file to compute heating and cooling loads for every time step.

Co-simulation refers to synchronizing two or more simulations in parallel. In our case, it is the synchronization of reading observations from EnergyPlus, and generating action for next time step from RLlib This turns out to be a challenge since RLlib and EnergyPlus runs on different processes. EnergyPlus provides support for BCVTB (Building Controls Virtual Test Bed), which is a software environment that allows users to couple different simulation programs for co-simulation. BCVTB uses Ptolemy server in the backend. This solution works, when one program is controlling one run of EnergyPlus, but with RLlib, it can spin multiple rollout workers, and each of them would need to start an EnergyPlus process. BCVTB based solution does not work in this scenario.

\begin{figure*}[ht]
	\centering
	\includegraphics[width=0.8\linewidth]{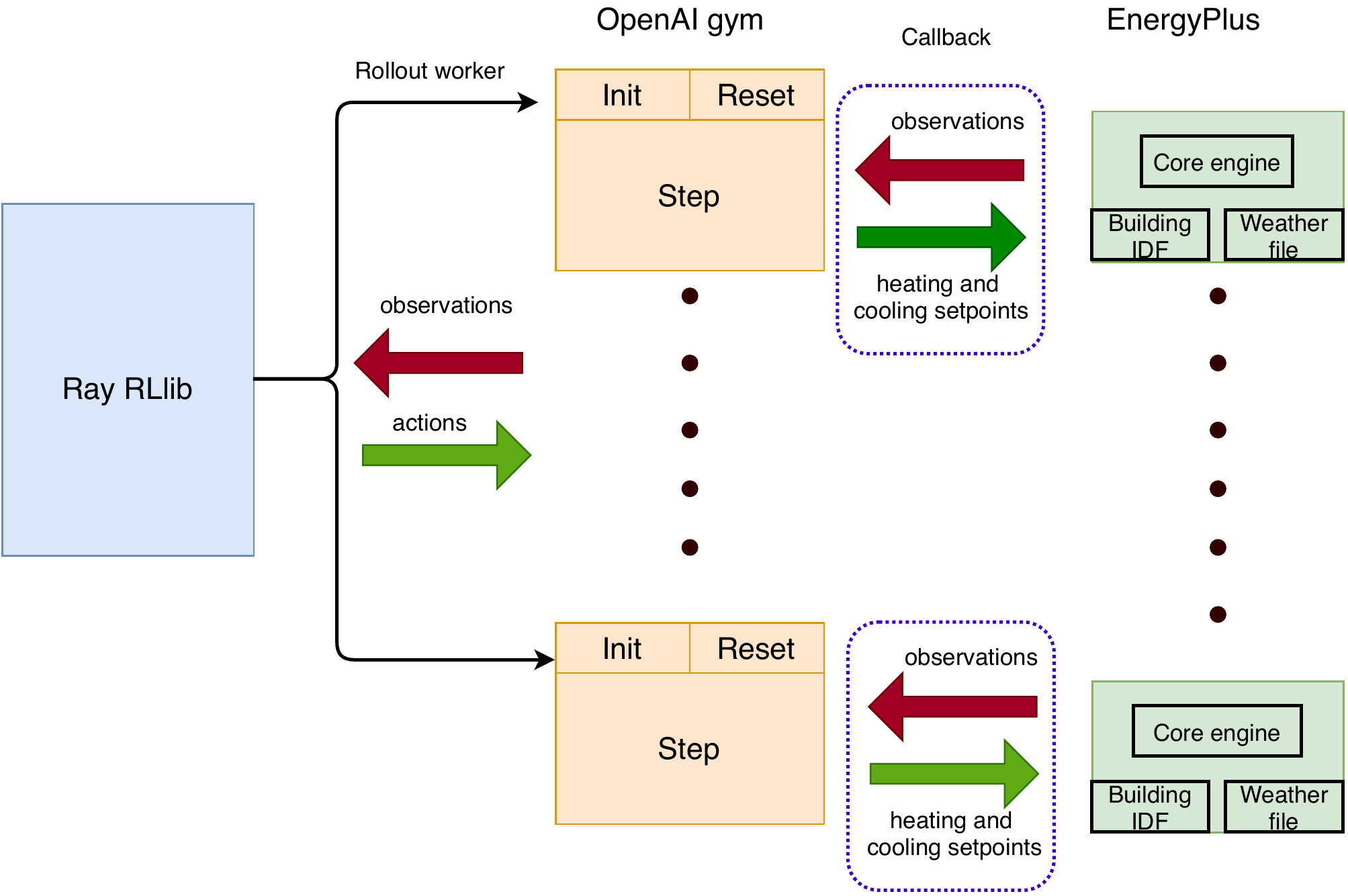}
	\caption{Distributed co-simulation framework involving Ray RLlib, OpenAI gym, and EnergyPlus}
	\label{fig:arch}	
	\Description{co-simulation framework}
\end{figure*}

Fortunately, EnergyPlus from version 9.3 onward provides\\ \texttt{pyenergyplus}, a Python library that contains callbacks to various states within EnergyPlus. Using this callbacks, we can intercept EnergyPlus, gather variables of interest, and pause the simulator until the necessary calculation needs to be done by RLlib. In order for RLlib to gather outputs from EnergyPlus and compute next action based on reward, we utilize the callback function \texttt{callback\_end\_zone\_timestep\_after\_zone\_reporting},\\ which returns to our Gym class at the end of every zone timestep after generating outputs. In this function, we collect outputs and put it in a queue, that will be picked by Gym step function, and compute the corresponding next action. Queues act as synchronization mechanism between our Gym environment and EnergyPlus. The framework is depicted in Figure~\ref{fig:arch}.

In order to change the setpoints on every timestep, we add following constant schedules in building IDF:
\begin{Verbatim}[frame=single]
Schedule:Constant,
  CLGSETP_SCH_Perimeter_top_ZN_4,   !- Name
  Temperature,         !- Schedule Type Limits Name
  20.0;                             !- Hourly Value
\end{Verbatim}
Through EnergyPlus callback, we modify this setpoint to desired setpoint, which will be held constant for the given timestep.

\section{Deep Reinforcement Learning for Building HVAC Control}
\label{sec:drl}

Reinforcement learning (RL) is a branch of machine learning that is specialized for solving control problems. Unlike classical control problems, where action is usually based on the immediate feedback, in many RL problems, the optimal policy could depend on delayed feedback, or need to discount current feedback. A RL problem primarily consists of following components:
\begin{itemize}
	\item \textbf{State}: State is the mathematical description of the environment. For example, in the case of HVAC control of a building, state would represent the current temperatures and humidity in various zones of the building, outdoor environment temperature and humidity, heating and cooling loads, etc.
	\item \textbf{Action}: Action represents the control taken on the environment. In the example of HVAC control, action could be setting the thermostat setpoints for each zone, fan speeds, etc.
	\item \textbf{Agent}: The job of an agent is to compute an optimal action for a given state, under an optimal policy.
	\item \textbf{Environment}: Environment is the super set, where one or more actions, result in a state transition with associated observations. Each state transition has an associated transition probability that is dependent on the current state and the action taken. Depending on the objective, a reward function is used to predict immediate rewards for an action under a state.
\end{itemize}

The goal of an RL agent is to compute an optimal policy given a sequence of actions and subsequent observations. There are primarily two ways of achieving this goal:
\begin{itemize}
	\item \textbf{Model-based RL}: Model-based RL is used in environments where we know the functions for transition probability and rewards are known. In this case, we can use policy or value iteration to find the optimal policy.
	\item \textbf{Model-free RL}: In most environments we don't know the exact characteristics of the environment. In that case, the controller needs to determine the optimal policy without modeling the environment. Policy gradient, value-based, and actor-critic are some of the common approaches used in model-free RL. We use model-free approach in this work. 
\end{itemize}

\subsection{DRL algorithms}
\label{sec:drl_algos}

Many algorithms for DRL have been developed over last few years. They can be divided based on several factors, e.g. whether the algorithm is improving same policy - on-policy, or continuously exploring new policies - off-policy; whether the algorithms support continuous or discrete control; if the algorithms are gradient based or derivative free. We will discuss some of the popular DRL algorithms in this section that are also supported in RLlib. 

PPO (Proximal Policy Optimization)~\cite{schulman:arxiv17-ppo} is a popular on-policy algorithm, which improves upon TRPO (Trust Region Policy Optimization)~\cite{schulman:arxiv17-trpo} by avoiding Kullback–Leibler divergence as a hard constraint, instead add the constraint as a penalty, which simplifies the computation. DDPG~\cite{lillicrap:arxiv19} is another model-free off-policy algorithm designed for learning continuous actions. DDPG uses Experience Replay and slow-learning target networks from DQN (Deep Q-Network)~\cite{mnih:arxiv13}, and it is based on DPG (Deterministic Policy Gradient)~\cite{silver:icml14}. 

Asynchronous Actor-Critic (A2C) and Advantage Actor-Critic (A3C)~\cite{mnih:arxiv16} methods updates the gradients asynchronously, which improves training times. Soft Actor Critic (SAC)~\cite{haarnoja:arxiv19} is an off-policy actor-critic DRL algorithm based on the maximum entropy reinforcement learning frame-work, where the actor aims to maximize expected reward while also maximizing entropy. Ape-X~\cite{horgan:arxiv18} provides a distributed architecture for DRL. When combined with DDPG, it can be used to scale across multiple instances.

\subsection{Multi-zone HVAC Control}

A building usually has multiple zones that could be spread over multiple floors. In a normal operation, these setpoints are set at a constant value. Since the range of desired temperature is a range, the zonal setpoints need not be set to a constant setpoint, but can be altered to match with external weather to reduce energy consumption. The setpoints for all zones could be set the same value, or they could be set independently to accommodate for differential temperatures between zones, e.g. core zones are less impacted by external weather than the perimeter zones. We will explore these options in detail in Section~\ref{sec:exp-multi-zone}.

\subsection{Multi-agent}

In a multi-agent setting, multiple agents are making control decision on an environment. There are two ways of setting up a multi-agent problem:
\begin{itemize}
	\item \textbf{Cooperative}: In this setting, agents share their observations, and might have a common reward. The idea behind this is to maximize rewards for all agents.
	\item \textbf{Competitive}: Here agents do not share their observation. They only try to maximize their individual rewards.
\end{itemize}

For our experiments, we consider cooperative type multi-agents. Specifically, we create two agents, one to control heating setpoint and another to control cooling setpoint. The intuition behind this is that each agent can create a separate model, so that agent responsible for setting heating setpoints can optimize for colder months, while the other agent can optimize for warmer months.
\section{Experiments}
\label{sec:experiments}

\subsection{Setup}

The following are some of the important settings we use in our experiments.

\begin{itemize}
	\item \textbf{EnergyPlus}: v9.3.0
	\item \textbf{Building}: DOE Commercial Reference Building Medium office, new construction 90.1-2004 (RefBldgMediumOfficeNew2004\_Chicago.idf~\footnote{\url{https://bcl.nrel.gov/node/85021}}).
	\item \textbf{Weather file}: Toronto, Canada (CAN\_ON\_Toronto.716240\_\\CWEC.epw~\footnote{\url{https://energyplus.net/weather-location/north_and_central_america_wmo_region_4/CAN/ON/CAN_ON_Toronto.716240_CWEC}}). We chose Toronto location, as it has good difference in temperature values between summer and winter months.
	\item \textbf{Simulation days}: We simulate for an entire year, i.e. 365 days. For comparison, we also simulate for 30 days in the month of January and July in Section~\ref{sec:exp-sim-days}.
	\item \textbf{Zones considered}: The reference building has three floors and each floor is subdivided into multiple core (center) and perimeter zones. The details of the zones and their names are listed in the next paragraph. We consider all of these zones.
	\item \textbf{Controls}: Heating and cooling setpoints for all zones.
	\item \textbf{Reward}: We use the reward defined in Equation~\ref{eqn:obj}.
	\item \textbf{Algorithms}: APEX\_DDPG~\cite{lillicrap:arxiv19} and PPO~\cite{schulman:arxiv17-ppo}.
	\item \textbf{RL library}: Amazon SageMaker RL~footnote{\url{https://docs.aws.amazon.com/sagemaker/latest/dg/reinforcement-learning.html}} with Ray RLlib, version 1.0.
	\item \textbf{Deep learning framework}: PyTorch 1.8.1.
	\item \textbf{Python version}: 3.6.		
	\item \textbf{Simulation timestep}: 15 minutes.
	\item \textbf{Desired range of zone temperatures}: 20\degree C -- 25\degree C.
	\item \textbf{Heating setpoint range}: 15\degree C to 22\degree C.
	\item \textbf{Cooling setpoint range}: 22\degree C to 30\degree C.
	\item \textbf{Training instance}: ml.g4dn.16xlarge, which has 64 cores, 256 MB of RAM, and 1 NVIDIA T4 GPU.
\end{itemize}

The reference building has following 15 zones: \texttt{Core\_bottom}, \texttt{Core\_mid}, \texttt{Core\_top}, \texttt{Perimeter\_bot\_ZN\_1}, \texttt{Perimeter\_bot\_ZN\_2}, \texttt{Perimeter\_bot\_ZN\_3}, \texttt{Perimeter\_bot\_ZN\_4}, \texttt{Perimeter\_mid\_ZN\_1}, \texttt{Perimeter\_mid\_ZN\_2}, \texttt{Perimeter\_mid\_ZN\_3}, \texttt{Perimeter\_mid\_ZN\_4}, \texttt{Perimeter\_top\_ZN\_1}, \texttt{Perimeter\_top\_ZN\_2}, \texttt{Perimeter\_top\_ZN\_3}, and \texttt{Perimeter\_top\_ZN\_4}. Core refers to the center of a floor, perimeter refers to the zones on the periphery of a floor. `top', `mid', and `bottom' refers to the location of the floors.

In addition to zone temperatures, we use following observations:
\begin{itemize}
	\item \texttt{Air System Total Heating Energy}, which is heat added to the air loop (sum of all components) in Joules. Similarly, \texttt{Air System Total Cooling Energy} is heat removed from the air loop. These variables are collected for each VAV (variable air volume) 1, 2, and 3. Their sum is represented as $P^h_z$ and $P^c_z$  in Table~\ref{tab:nomenclature}, respectively. VAV is a type of HVAC system, which unlike constant air volume systems can vary the airflow at a constant temperature.
	\item \texttt{Site Outdoor Air Wetbulb Temperature} and \texttt{Site\\ Outdoor Air Drybulb Temperature}. Dry-bulb temperature is the temperature that measures air temperature without the effect of any moisture. On the other hand, wet-bulb temperature is the lowest temperature that can be reached under current ambient conditions by the evaporation of water only. Since wet-bulb temperature accounts for relative humidity, we include that as one of the observables.
\end{itemize}

In the following sections, we will explore the effect of various configurations on the reward optimization and training times.

\subsection{Effect of $\alpha$ and Seasons on Reward}
\label{sec:exp-sim-days}

\begin{figure}[ht]
	\centering	
	\begin{subfigure}[ht]{\linewidth}
		\centering
		\includegraphics[width=\textwidth]{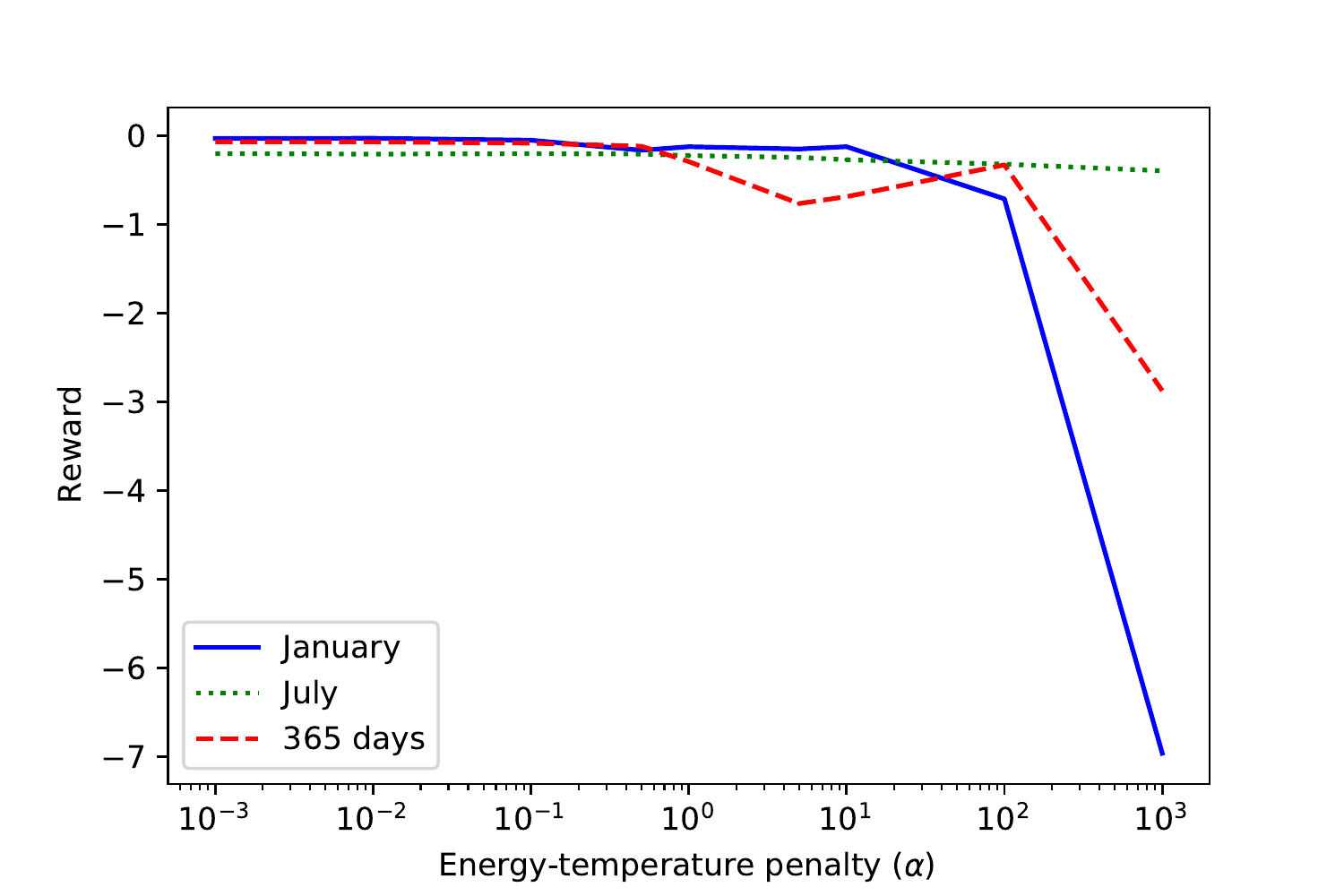}
		\caption{Plot of final reward.}
		\label{fig:result-sim-days-reward}
		\Description{RL: Impact of weather - reward}		
	\end{subfigure}
	\begin{subfigure}[ht]{\linewidth}
		\centering
		\includegraphics[width=\textwidth]{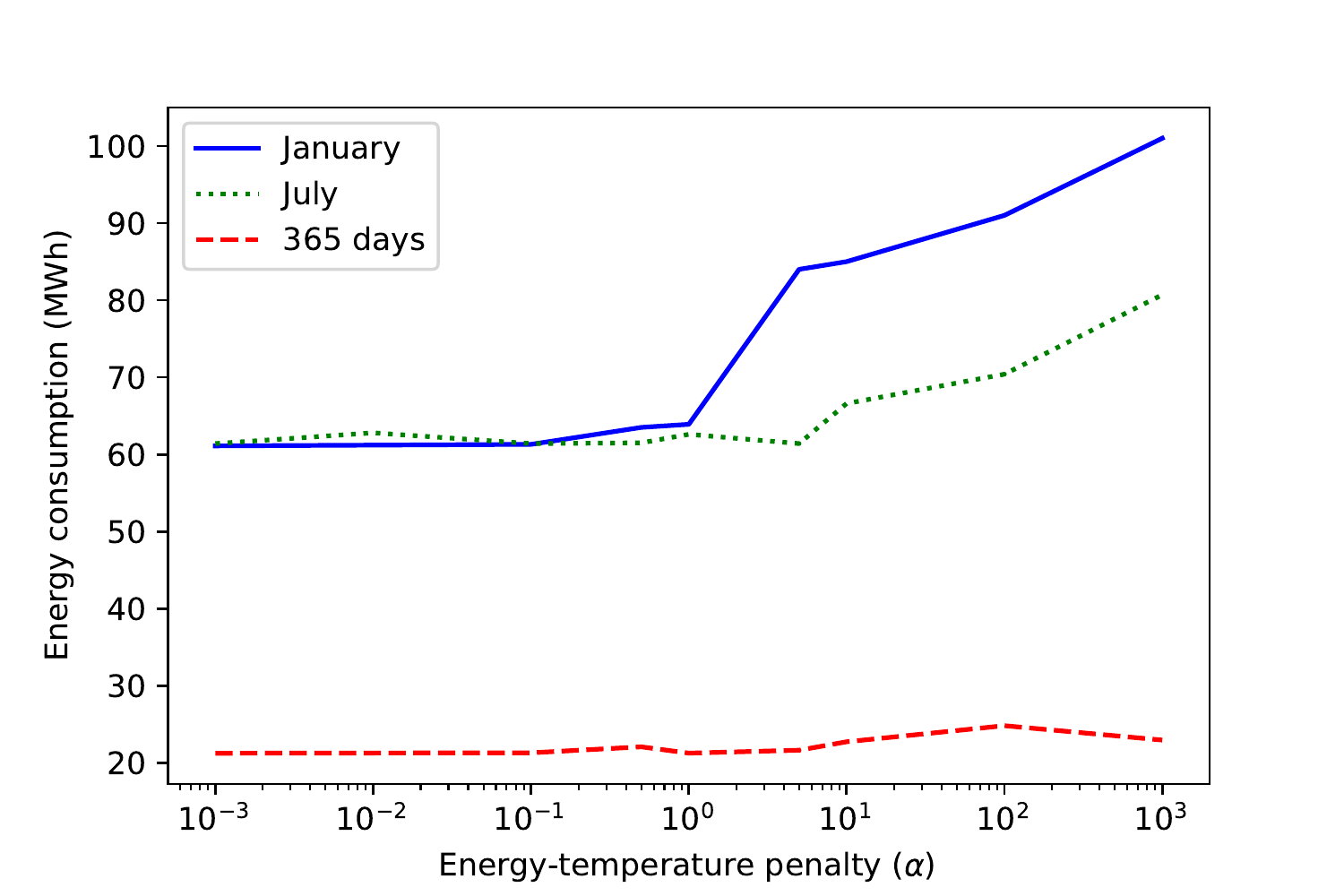}
		\caption{Plot of energy consumption. The energy consumption of 365 days simulation is normalized for 30 days for comparison reasons.}
		\label{fig:result-sim-days-energy}
		\Description{RL: Impact of weather - energy}		
	\end{subfigure}
	\begin{subfigure}[ht]{\linewidth}
		\centering
		\includegraphics[width=\textwidth]{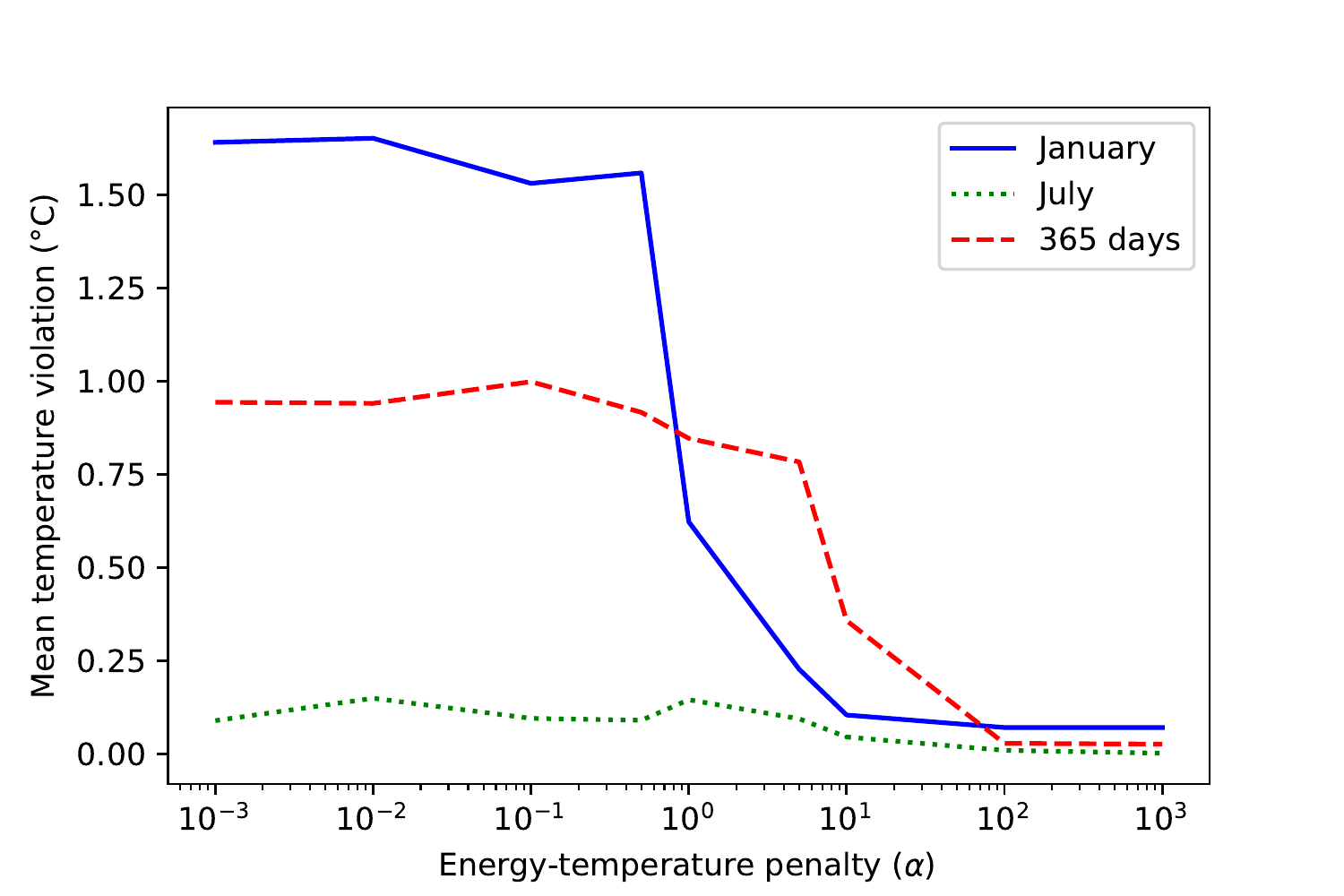}
		\caption{Plot of mean temperature violation.}
		\label{fig:result-sim-days-temp}
		\Description{RL: Impact of weather - temperature}		
	\end{subfigure}
	\caption{Plot showing the effect of number of simulation days and season on the reward for various $\alpha$.}
	\label{fig:result-sim-days}
	\Description{RL: Simulation days vs reward}	
\end{figure}

As described earlier, $\alpha$ controls if we are optimizing more for energy savings vs temperature comfort. The final reward of a simulation depends not only on $\alpha$, but also on the number of simulation days, and the seasons captured in the simulation run. Since the rewards generated for every time step are accumulated towards final reward, we need to normalize the rewards for number of simulation days. Figure~\ref{fig:result-sim-days} shows the comparison of reward function, energy consumption, and mean violation of zone temperatures for various $\alpha$, when simulated for 30 days in the months of January and July, and also for entire year (365 days). January and July were chosen as these two months are representative of cold and warm months in a typical climate. The raw data points are provided in Table~\ref{tab:result-sim-days} in the Appendix~\ref{sec:appendix-sim-days}.

From the plots, we see that in general, energy consumption is higher in winter, as the differential between external temperature and the indoor temperature is higher in winter than in summer (for Toronto region). Similar reasoning also applies for mean temperature violation, where we see higher violation for January month than for July. Simulating for entire year tries to find a policy that works for both seasons, and ends up not being best in both seasons. It is for this reason, we explore using multi-agent policy (Section~\ref{sec:exp-multi-agent}), where one policy focuses on wamer months, while other on colder months.

\textbf{Comparison with baseline:} In addition to above experiment, we simulated the baseline experiment, where we set the heating setpoint to 20$\degree$C and cooling setpoint  to 22.5$\degree$C for all zones, as they are the minimum values to avoid temperature violation yet, and thus minimize energy consumption. The result was the total baseline energy is 517.41~MWh and mean temperature violation is  0.118~$\degree$C. Comparing with centralized policy, we see that RL algorithms at the minimum (for $\alpha = 0.001$) gives \textbf{75\%} reduction in energy consumption and mean temperature lower is by 0.116~$\degree$C.

\subsection{Multi-zone}
\label{sec:exp-multi-zone}

\begin{figure}[ht]
	\centering	
	\begin{subfigure}[ht]{\linewidth}
		\centering
		\includegraphics[width=\textwidth]{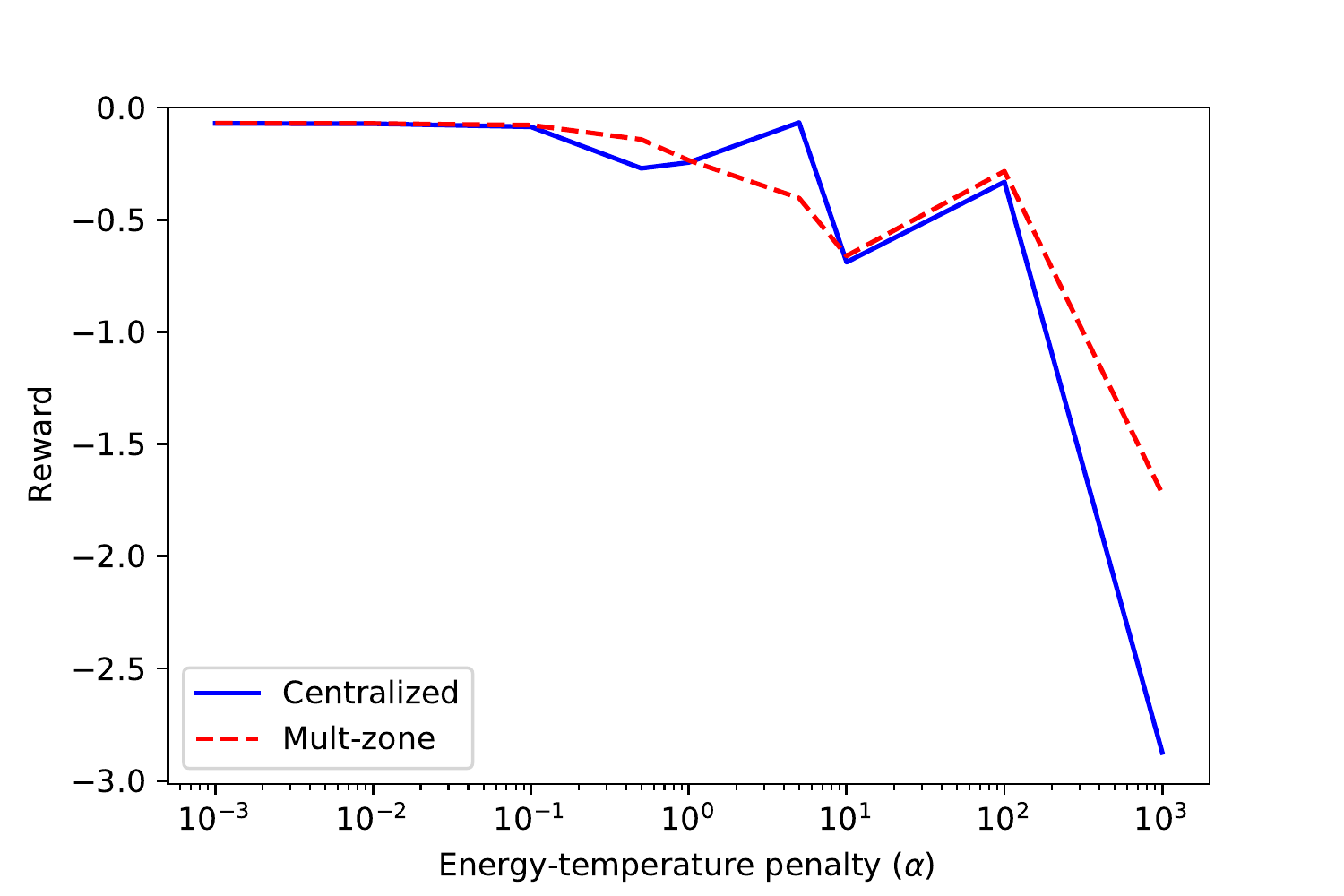}
		\caption{Plot of final reward.}
		\label{fig:result-multi-zone-reward}
		\Description{RL: Impact of weather - reward}		
	\end{subfigure}
	\begin{subfigure}[ht]{\linewidth}
		\centering
		\includegraphics[width=\textwidth]{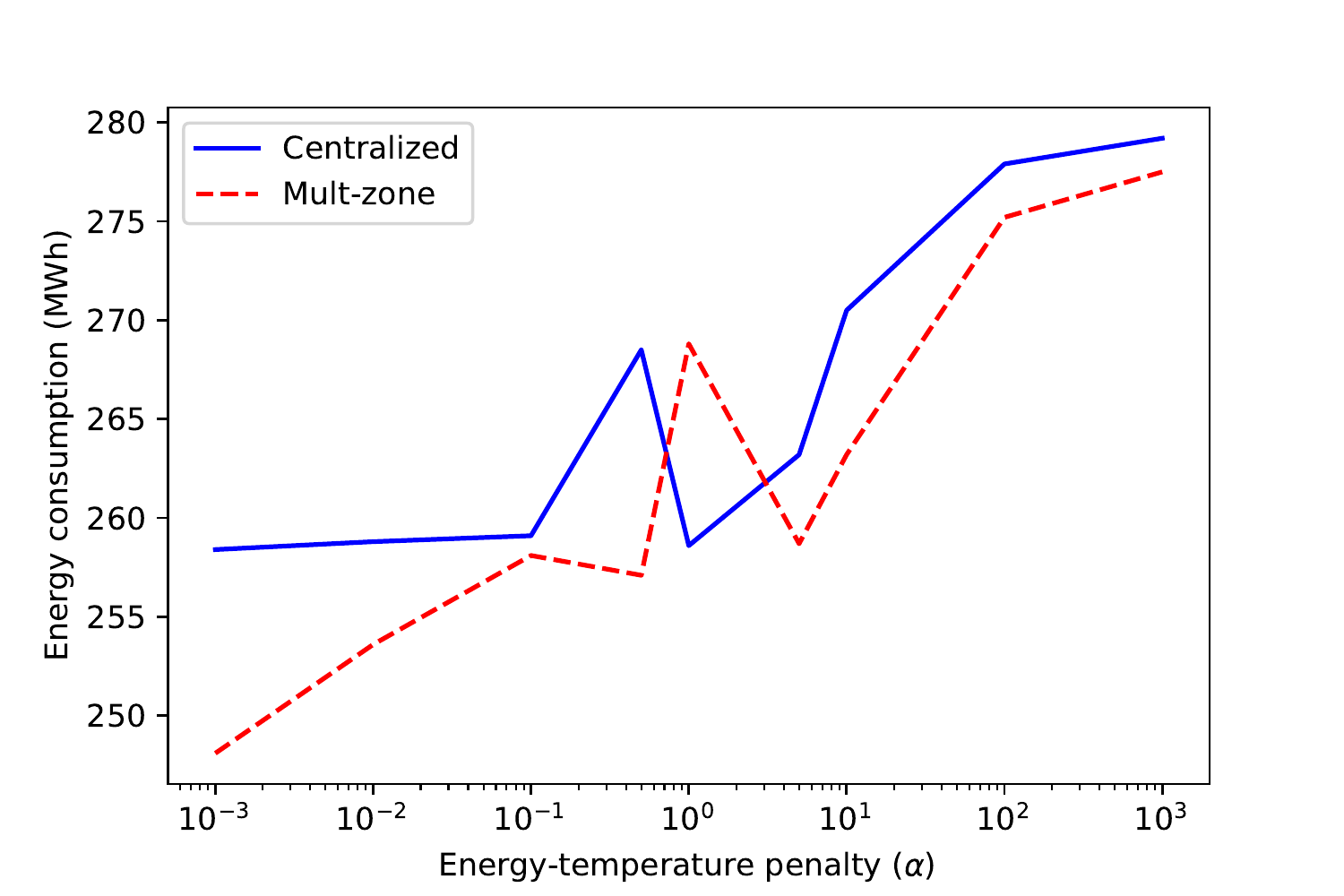}
		\caption{Plot of energy consumption.}
		\label{fig:result-multi-zone--energy}
		\Description{RL: Impact of weather - energy}		
	\end{subfigure}
	\begin{subfigure}[ht]{\linewidth}
		\centering
		\includegraphics[width=\textwidth]{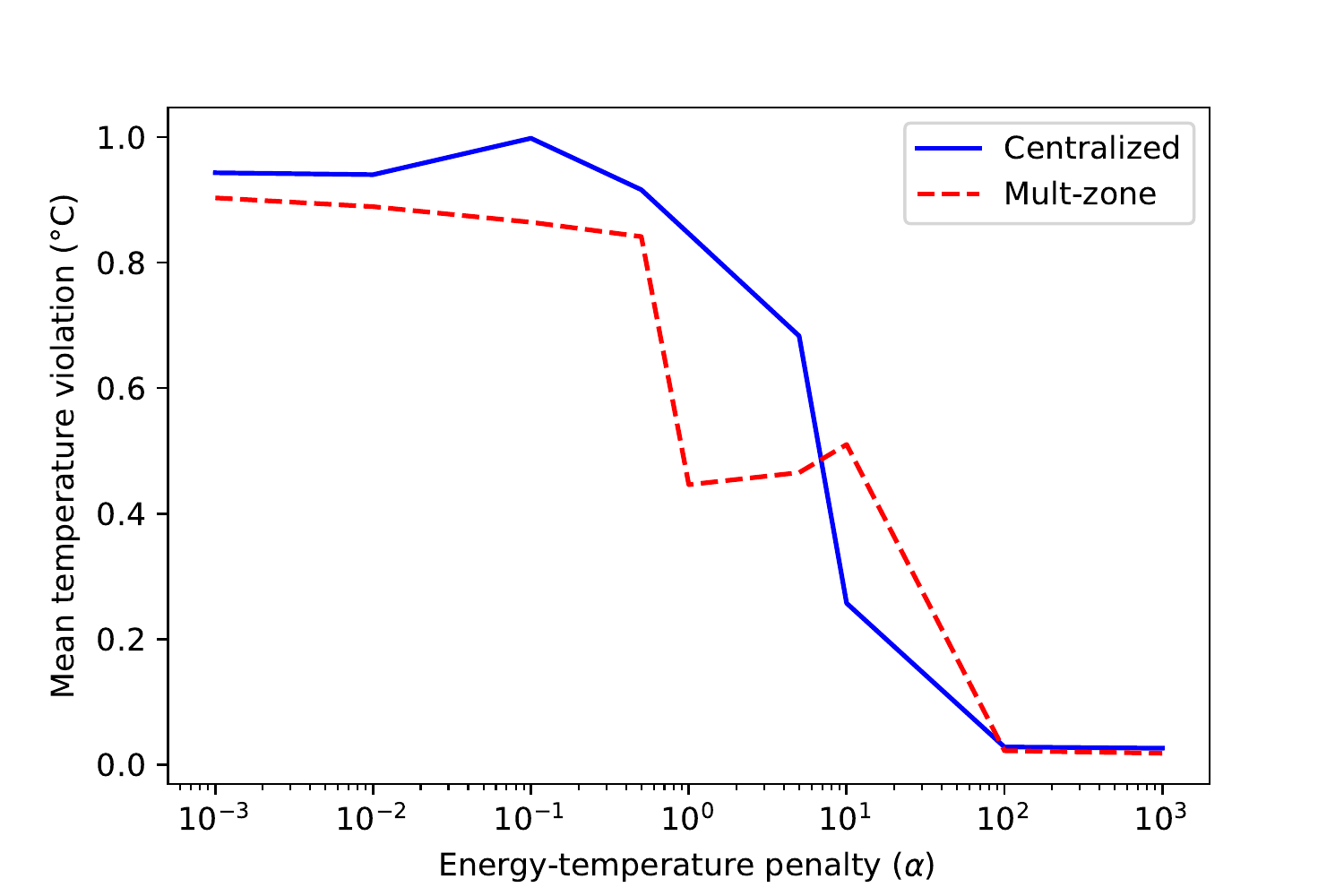}
		\caption{Plot of mean temperature violation.}
		\label{fig:result-multi-zone-temp}
		\Description{RL: Impact of weather - temperature}		
	\end{subfigure}
	\caption{Plot comparing multi-zone policy vs centralized policy for various $\alpha$.}
	\label{fig:result-multi-zone}
	\Description{RL: Simulation days vs reward}	
\end{figure}

There are two approaches in controlling HVAC of a building - one is centralized control, where every zone gets same heating and cooling setpoint at a given time, the other is individualized control for each zone. We evaluate the rewards for various $\alpha$, and energy consumption and mean temperature violation for above approaches. These observations are summarized in Figure~\ref{fig:result-multi-zone}. From the figure, we see that multi-zone approach does better in overall reward value, energy consumption, and also on mean temperature violation over the entire range of $\alpha$. For couple of datapoints, centralized policy seems to be doing better than multi-zone. This cross-over value is small and we believe it could be due to random nature of policy optimization. This agrees with our intuition that different zones in a building need different temperature setting as all zones don't have same thermal dynamics,e.g. a perimeter zone on south side will receive more sun than the core zones, and thus would need a different setpoint than the core zones. Please refer to Appendix~\ref{sec:appendix-multi-zone} for raw data used in the plots.

\subsection{Impact of Weather on HVAC Optimization}
\label{sec:exp-weather}

\begin{figure}[ht]
	\centering	
	\begin{subfigure}[ht]{\linewidth}
		\centering
		\includegraphics[width=\textwidth]{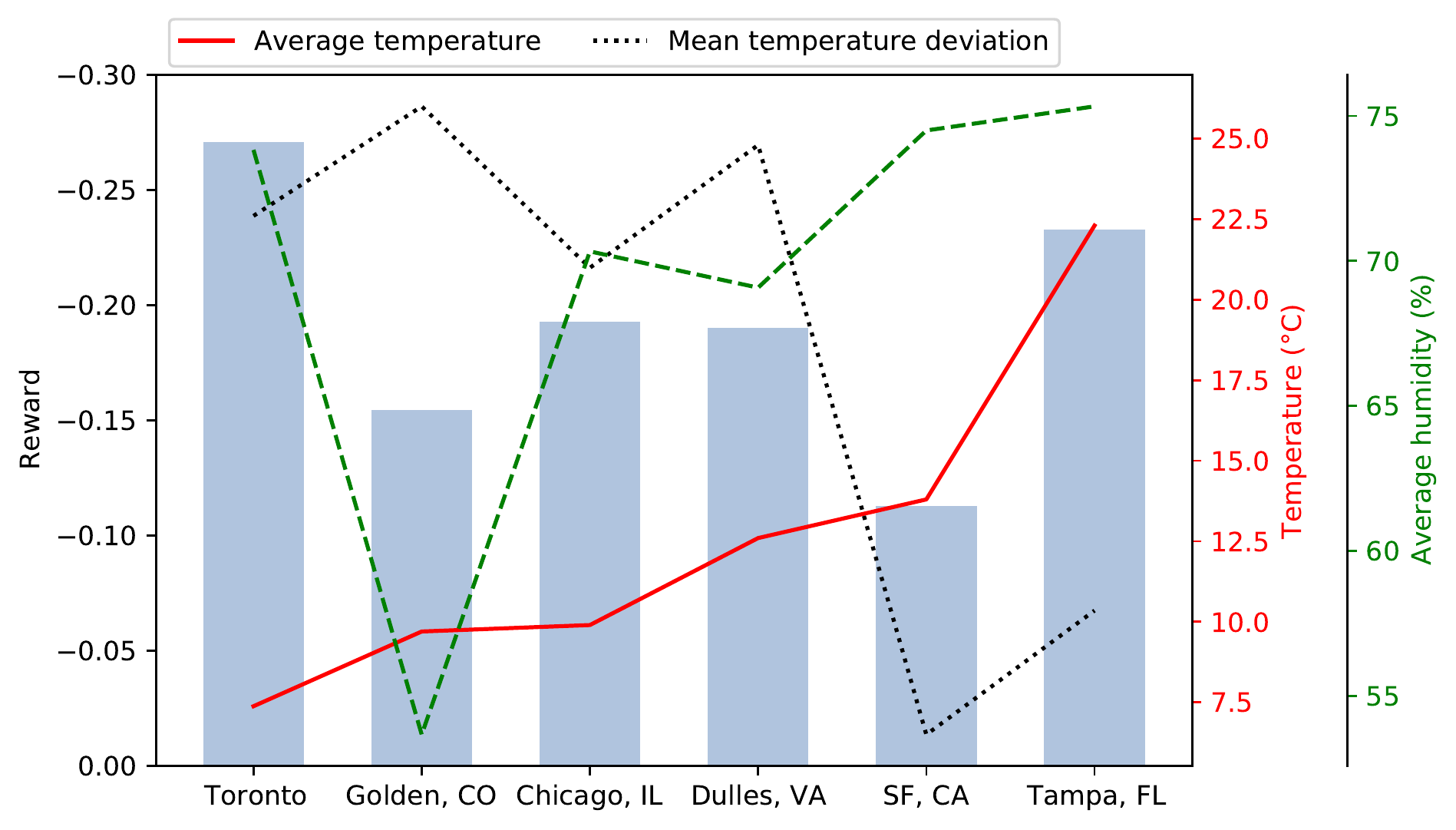}
		\caption{Plot of final reward.}
		\label{fig:result-weather-reward}
		\Description{RL: Impact of weather - reward}		
	\end{subfigure}
	\begin{subfigure}[ht]{\linewidth}
		\centering
		\includegraphics[width=\textwidth]{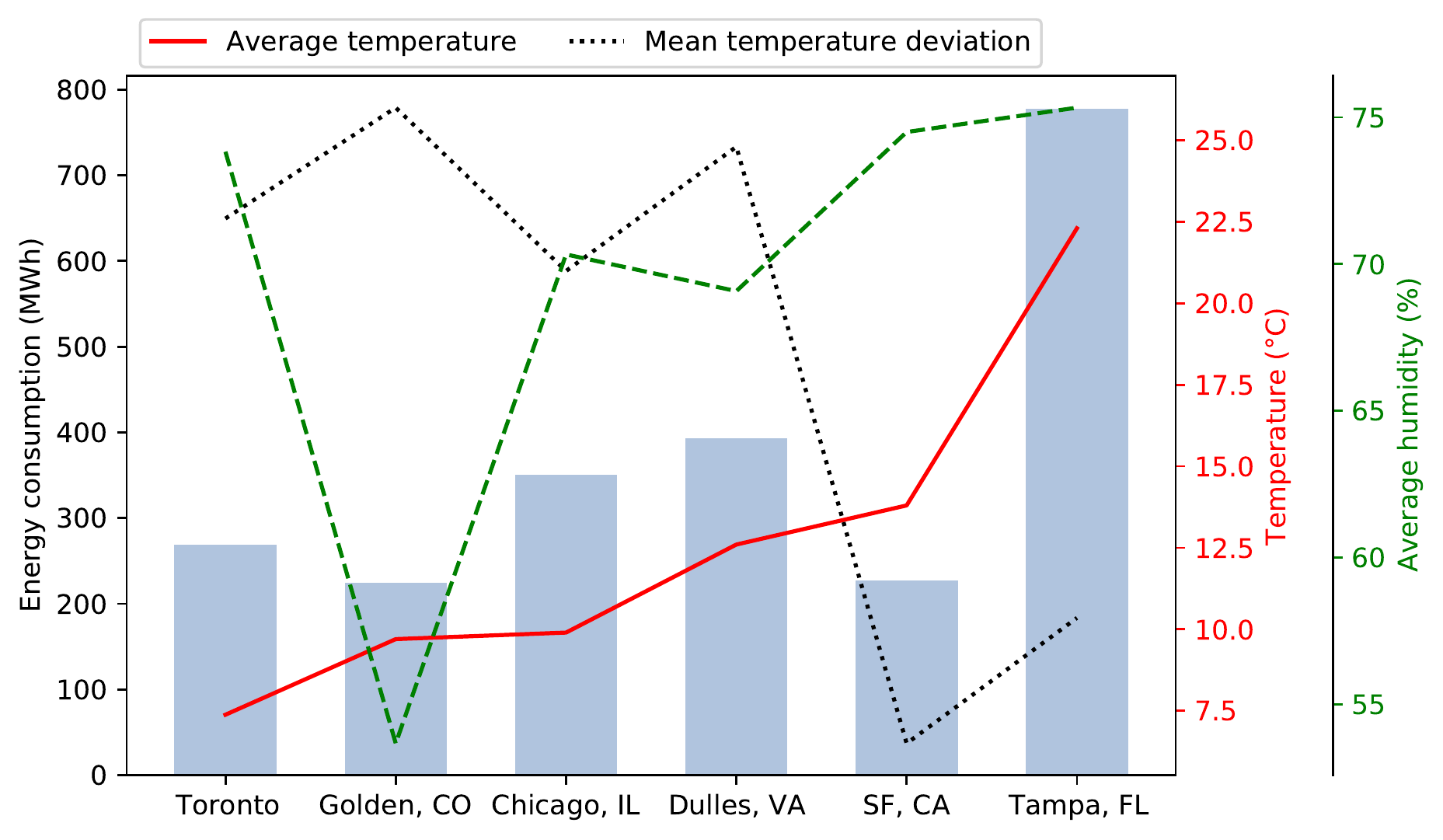}
		\caption{Plot of energy consumption.}
		\label{fig:result-weather-energy}
		\Description{RL: Impact of weather - energy}				
	\end{subfigure}
	\begin{subfigure}[ht]{\linewidth}
		\centering
		\includegraphics[width=\textwidth]{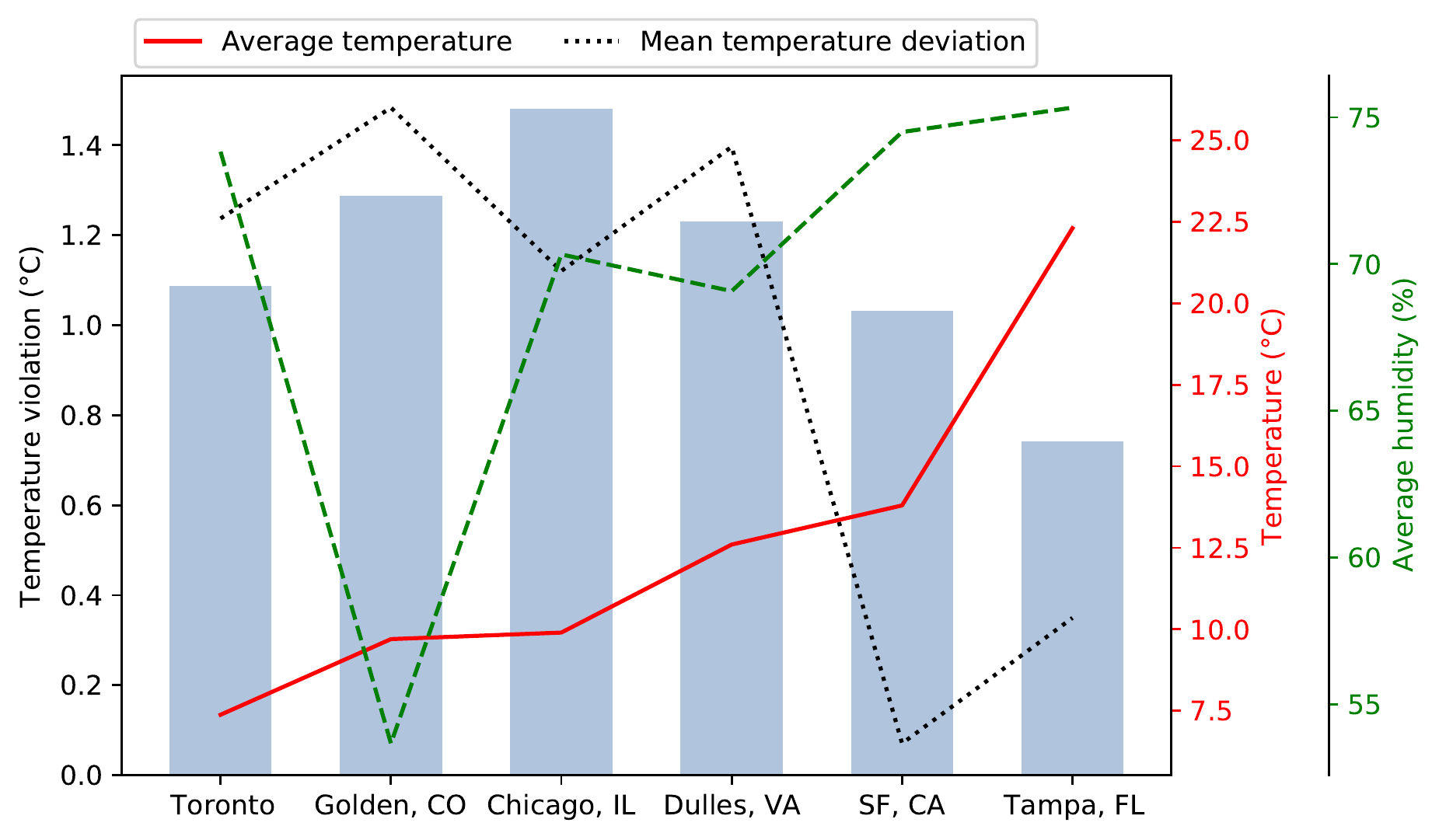}
		\caption{Plot of mean temperature violation.}
		\label{fig:result-weather-temp}
		\Description{RL: Impact of weather - temperature}				
	\end{subfigure}
	\caption{Plot of impact of external weather on HVAC optimization. Each plot contains metrics for 6 cities along with plot of average temperature, average humidity, and mean temperature deviation from summer and winter months. $\alpha$ was set to 1.}
	\label{fig:result-weather}
	\Description{RL: Impact of weather}	
\end{figure}

In this section, we consider the impact of external weather on HVAC optimization of a building. We use weather from 6 locations and perform multi-zone optimization as described in the previous section. Figure~\ref{fig:result-weather} describes the plots of reward, energy consumption, and mean temperature violation for different locations. The average temperature and humidity for entire year for each location, along with temperature difference between summer and winter months are overlay on the plots for better understanding. The details of data and full names of location can be found in Appendix~\ref{sec:appendix-weather}.

From the plots, we see that the average external temperature plays a big role in reducing energy consumption, which also impact the reward. Both low and high average temperatures decreases reward (see Toronto and Tampa, FL), as that would require more energy to keep indoor temperature in the comfort zone. Note that the x-axis in Figure~\ref{fig:result-weather-reward} is inverted. The lower the bar, better the rewad value. In general, it takes more energy to cool than to heat, as we can see from Figure~\ref{fig:result-weather-energy}, where higher average temperatures in Tampa, FL requires more than double the energy consumption for a building to cool than for a building in Toronto, where the temperatures on average are much cooler.. 

Mean difference in summer and winter temperatures also plays a role, as it represents the differential work, the HVAC systems need to do to keep indoor temperatures within comfort zone, across different seasons. Similarly, higher humidity keeps the outdoor temperatures relatively constant, which reduces mean temperature violation in zones, as see with San Francisco location.
 
\subsection{Multi-agent}
\label{sec:exp-multi-agent}

\begin{figure}[ht]
	\centering	
	\begin{subfigure}[ht]{\linewidth}
		\centering
		\includegraphics[width=\textwidth]{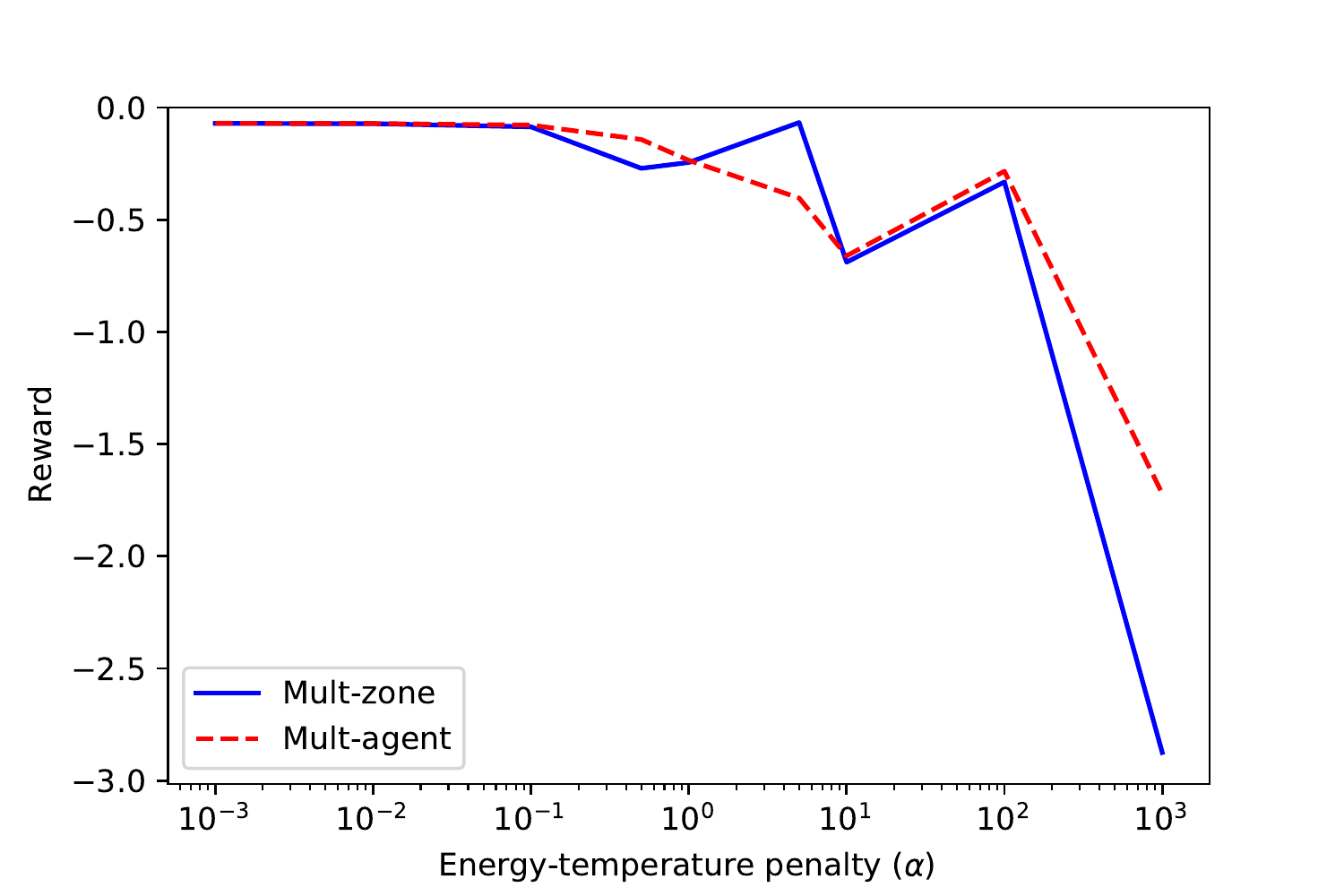}
		\caption{Plot of final reward. Note the x-axis is inverted. The smaller the length of the bar, better the reward.}
		\label{fig:result-multi-agent-reward}
		\Description{RL: multi-agent - reward}		
	\end{subfigure}
	\begin{subfigure}[ht]{\linewidth}
		\centering
		\includegraphics[width=\textwidth]{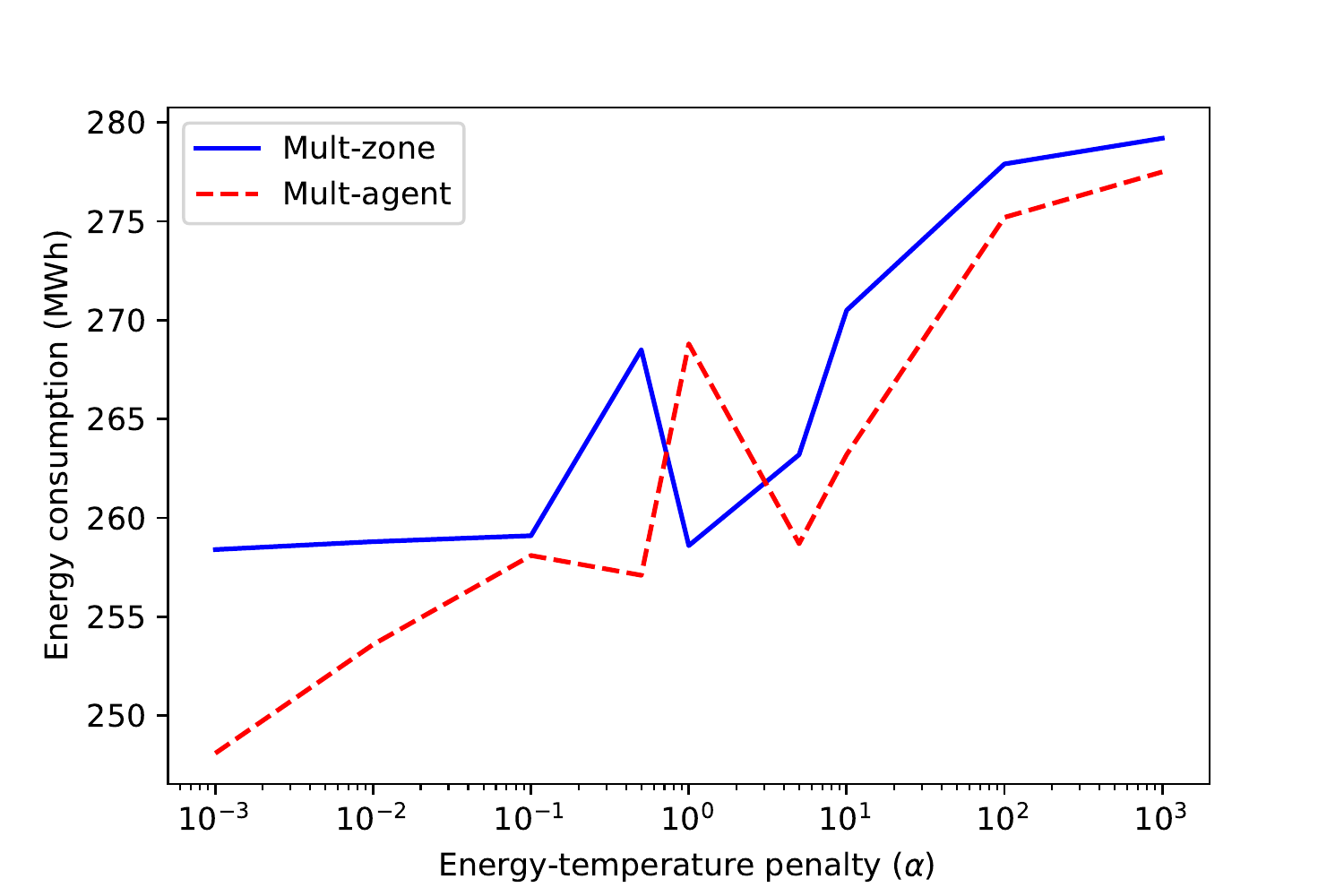}
		\caption{Plot of energy consumption.}
		\label{fig:result-multi-agent-energy}
		\Description{RL: multi-agent - energy}		
	\end{subfigure}
	\begin{subfigure}[ht]{\linewidth}
		\centering
		\includegraphics[width=\textwidth]{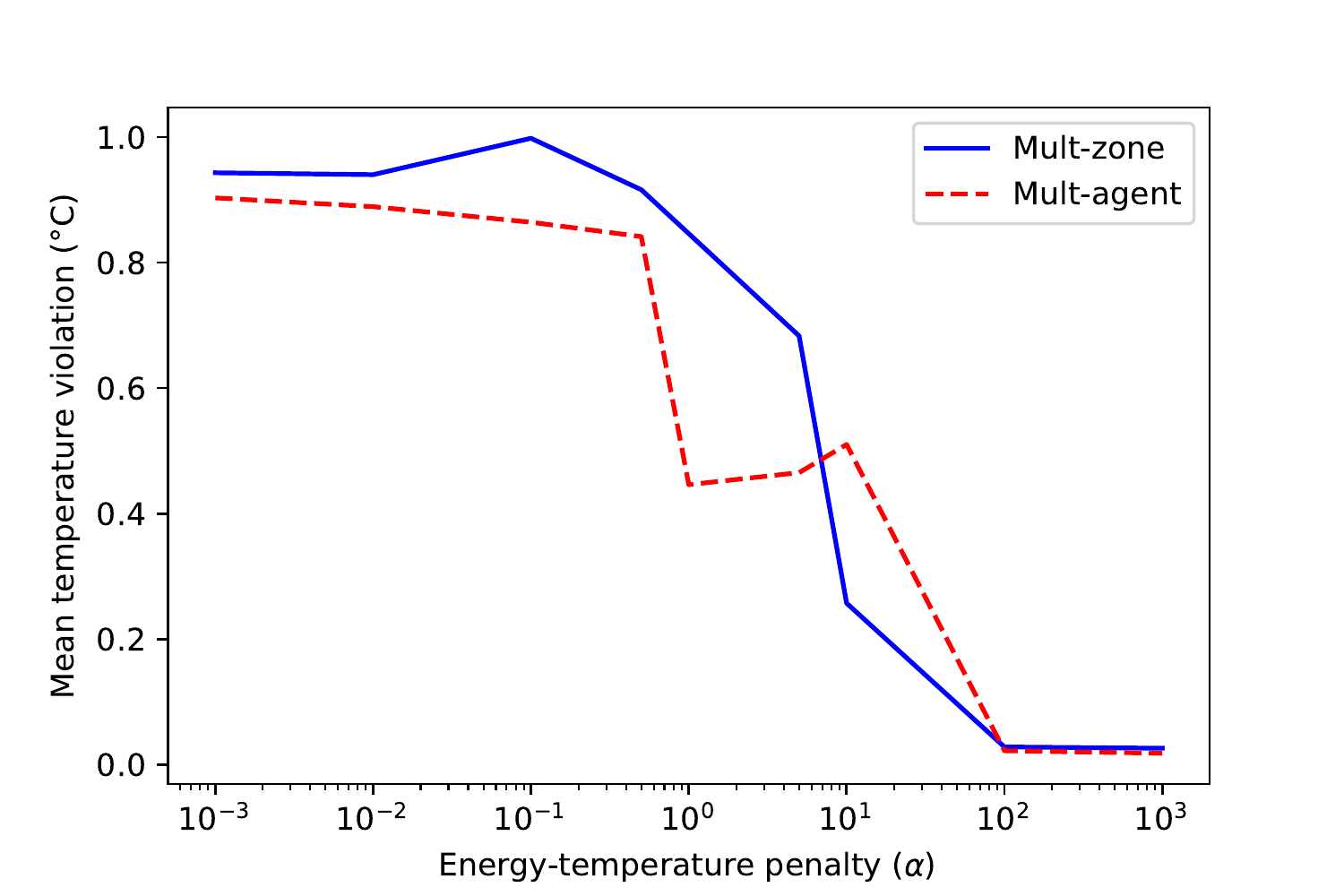}
		\caption{Plot of mean temperature violation.}
		\label{fig:result-multi-agent-temp}
		\Description{RL: multi-agent - temperature}
	\end{subfigure}
	\caption{Plot comparing multi-zone with multi-agent policies for various $\alpha$.}
	\label{fig:result-multi-agent}
	\Description{RL: Impact of weather}	
\end{figure}

Next, we explored setting up multiple agents to control HVAC. The idea behind multi-agent control is that multiple agents may better optimize reward, when the controls are not homogeneous. For example, in HVAC use case, we have two sets of controls - heating setpoint and cooling setpoint. Heating setpoint is used during winter months to increase indoor temperature, while cooling setpoint is used in summer months to reduce indoor temperature. A single agent will model the environment using a single neural network and uses same weights to decide next heating setpoint and cooling setpoint controls, which may not be optimal. With multi-agent, each agent (heating/cooling setpoint) will optimize individual network for the control they are responsible. Figure~\ref{fig:result-multi-agent} compares multi-agent policy with multi-zone policy we discussed earlier. From the figure, we notice that multi-agent does better than multi-zone in general across various $\alpha$.

\subsection{Comparison of RL Algorithms}

\begin{figure}[ht]
	\centering
	\includegraphics[width=\linewidth]{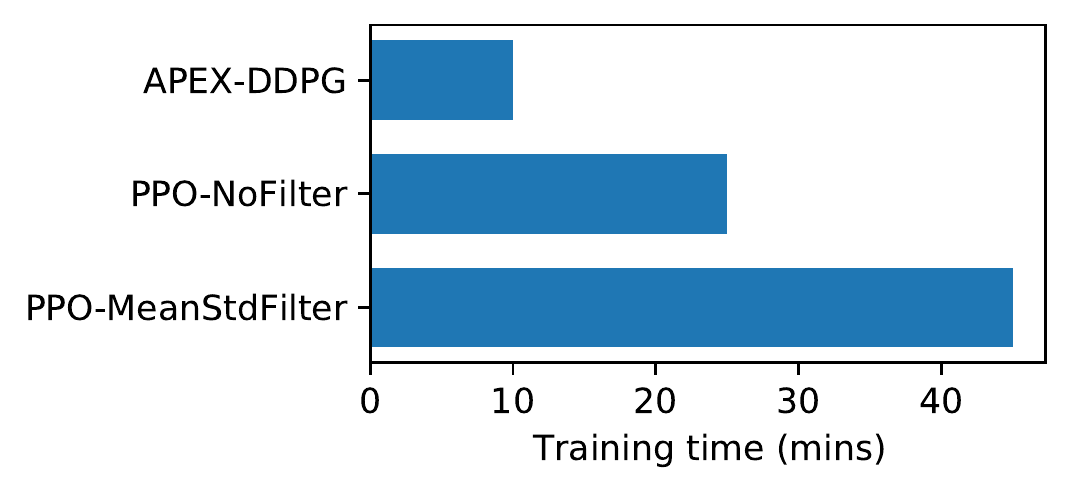}
	\caption{Plot comparing training times for PPO (NoFilter and MeanStdFilter observation filters) and APEX-DDPG algorithms.}
	\label{fig:result-algorithms}	
	\Description{RL: Simulation days vs reward}
\end{figure}

In this experiment, we compare two RL algorithms PPO and APEX-DDPG that were described in Section~\ref{sec:drl_algos}. Table~\ref{tab:result-algo-compare} and Figure~\ref{fig:result-algorithms} summarizes the comparison of different algorithms mentioned above. Rewards from PPO (for different observation filters) and APEX-DDPG are close. In terms of training times, APEX-DDPG has the lowest training time (\textbf{60\%} lower than PPO) and the reward converges in 84\% less iterations. \texttt{MeanStdFilter}, which keeps track of running mean, took longer to converge and with no significant improvement in reward.

\begin{table}[ht]
	\caption{Table comparing RL algorithms along with different configuration values, in terms of reward value and training time .}
	\label{tab:result-algo-compare}
	\begin{tabular}{cp{0.7in}cp{0.5in}p{0.4in}}
		\toprule
		Algorithm & Observation filter & Reward & Iteration count & Training time (s)\\
		\midrule		
		\multirow{2}{*}{PPO} & NoFilter & -293.19 & 104 & 25\\
		& MeanStdFilter & -292.73 & 178	& 45\\			
		APEX-DDPG & -- & -293.02 & 17 & 10\\			
		\bottomrule
	\end{tabular}
\end{table}

\subsection{Training Cost Optimization}

In the final experiment, we run our HVAC optimization on different AWS instances as listed in Table~\ref{tab:result-training-cost}. The table lists GPU types if the instances have a GPU, number of CPU cores, the training time, and the resulting cost. From the table, we can infer that having many cores is very helpful to rollout many RLlib workers, at the same time having a low cost GPU is useful to train the model in less time. For our experiments, we found ml.g4dn.16xlarge turns to be the low cost.

\begin{table}[ht]
	\caption{Table comparing convergence times on different instances on Amazon SageMaker and their respective costs.}
	\label{tab:result-training-cost}
	\begin{tabular}{llp{0.4in}p{0.4in}p{0.4in}}
		\toprule
		Instance type & GPU type & CPU cores & Training time (mins) & Training cost (\$)\\
		\midrule
		ml.g4dn.16xlarge & T4 & 64 & 15 & 1.36\\
		ml.p3.2xlarge & V100 & 8 & 40 & 2.55\\
		ml.p2.xlarge & K80 & 4 & 135	& 2.53\\
		ml.c5.18xlarge	& -- & 72 & 35 & 2.14\\
		ml.m5.24xlarge	& -- & 96 & 50 & 4.6\\
		\bottomrule
	\end{tabular}
\end{table}

\section{Conclusion}
\label{sec:conclusion}

In this paper, we have presented a scalable framework for optimizing HVAC on a commercial building using deep reinforcement learning. We have conducted experiments, showing the impact of simulation days, weather, energy-temperature penalty coefficient on the reward. We have also shown how to set up multi-zone and multi-agent control in this framework. We believe this framework will ease the adoption of DRL in HVAC optimization with researchers and practitioners, and spur further innovation.

%%
%% The next two lines define the bibliography style to be used, and
%% the bibliography file.
%\nocite{*}
\bibliographystyle{ACM-Reference-Format}
\bibliography{references}

%%
%% If your work has an appendix, this is the place to put it.
%\appendix

\appendix
\begin{table*}[htbp]
	\caption{Table with results on simulation of effect of energy-temperature penalty coefficient ($\alpha$), duration of simulation, and the month of simulation on reward value.}
	\label{tab:result-sim-days}
	\begin{tabular}{c|ccc|ccc|ccc}
		\toprule
		\multirow{2}{*}{$\alpha$} & \multicolumn{3}{c}{Reward} & \multicolumn{3}{c}{Energy consumption(MWh)} & \multicolumn{3}{c}{Mean temperature violation ($\degree$C)}\\
		& January (30) & July (30) & Year (365) & January (30) & July (30) & Year (365) & January (30) & July (30) & Year (365)\\
		\midrule		
		0.001 & -0.034 & -0.2045 & -0.0709 & 61.1 & 61.4 & 21.2 & 1.64 & 0.089 & 0.943\\
		0.01 & -0.0313 & -0.2095 & -0.0721 & 61.2 & 61.4 & 21.3 & 1.651 & 0.149 & 0.94\\
		0.1 & -0.0532 & -0.2055& -0.0861 & 61.3 & 62.8 & 21.3 & 1.53 & 0.095 & 0.998\\
		0.5 & -0.166 & -0.209	& -0.121 & 63.5 & 61.4 & 22.1 & 1.558 & 0.0898 & 0.916\\
		1 & -0.1267 & -0.2267 & -0.2943 & 63.9 & 61.5 & 21.3 & 0.622 & 0.145 & 0.846\\
		5 & -0.1523 & -0.2468 & -0.767 & 84 & 62.6 & 21.6 & 0.227 & 0.094 & 0.783\\
		10 & -0.1263 & -0.2733	& -0.6893 & 85 & 22.7 & 276.5 & 0.104 & 0.045 &  0.3571\\
		100 & -0.714 & -0.323	& -0.3318 & 91 & 24.8 & 301.9 & 0.071 & 0.0092 & 0.028\\
		1000 & -6.96 & -0.398 & -2.875 & 101 & 22.9 & 279.2 & 0.0704 & 0.0015 &  0.0261\\		
		\bottomrule
	\end{tabular}
\end{table*}

\section{Data used in  Plots Generation}

In this appendix, we will provided detailed data that was used to generate plots in Section~\ref{sec:experiments}.

\subsection{Effect of $\alpha$ and Seasons on Reward}
\label{sec:appendix-sim-days}

Table~\ref{tab:result-sim-days} provides the data used for generating Figure~\ref{fig:result-sim-days}.

\subsection{Comparison of Centralized vs Multi-zone Control}
\label{sec:appendix-multi-zone}

Table~\ref{tab:result-single-multi-zone} provides the data used for generating Figure~\ref{fig:result-multi-zone}.

\begin{table}[htbp]
	\caption{Table comparing centralized vs multi-zone control on reward value and training time as $\alpha$ is varied.}
	\label{tab:result-single-multi-zone}
	\begin{tabular}{ccp{0.6in}p{0.6in}c}
		\toprule
		Control type & $alpha$ & Energy consumption (MWh) & Mean temperature violation ($\degree$C) & Reward\\
		\midrule
		\multirow{9}{*}{Centralized} & 0.001 & 265.5 & 1.422 & -0.073\\
		& 0.01	& 266.2 & 1.287 & -0.075\\
		& 0.1 & 264.8 & 1.395 & -0.098\\
		& 0.5 & 264.8	& 1.395 & -0.298\\
		& 1	& 274.4 & 1.14 & -0.276\\
		& 5	& 265.4 & 0.568 & -0.851\\
		& 10 & 272.6 & 0.2131 & -0.4215\\
		& 100 & 276.9 & 0.497 & -5.492\\
		& 1000 & 289 & 0.02 & -2.053\\			
		\midrule								
		\multirow{9}{*}{Multi-zone} & 0.001	& 258.4 & 0.943 & -0.0709\\
		& 0.01	& 258.8 & 0.94	& -0.0721\\
		& 0.1 & 259.1 & 0.998	& -0.0861\\
		& 0.5 & 268.5 & 0.916	& -0.121\\
		& 1	& 258.6 & 0.846 & -0.2943\\
		& 5	& 263.2 & 0.783 & -0.767\\
		& 10 & 276.5 & 0.3571	& -0.6893\\
		& 100 & 301.9	& 0.028	& -0.3318\\
		& 1000 & 279.2	& 0.0261 & -2.875\\	
		\bottomrule
	\end{tabular}
\end{table}

\subsection{Impact of Weather on HVAC Optimization}
\label{sec:appendix-weather}

Table~\ref{tab:result-weather} provides the data used for generating Figure~\ref{fig:result-weather}. The following are the full names of the location found in NREL website.
\begin{itemize}
	\item CAN\_ON\_Toronto.716240\_CWEC
	\item USA\_FL\_Tampa.Intl.AP.722110\_TMY3
	\item USA\_CA\_San.Francisco.Intl.AP.724940\_TMY3
	\item USA\_CO\_Golden-NREL.724666\_TMY3
	\item USA\_IL\_Chicago-OHare.Intl.AP.725300\_TMY3
	\item USA\_VA\_Sterling-Washington.Dulles.Intl.AP.724030\_TMY3
\end{itemize}

\begin{table}[htbp]
	\caption{Table comparing centralized vs multi-zone control on reward value and training time as $\alpha$ is varied.}
	\label{tab:result-weather}
	\begin{tabular}{p{0.4in}p{0.25in}p{0.4in}p{0.4in}p{0.3in}p{0.35in}p{0.4in}}
		\toprule
		Location & Energy & Mean temperature violation ($\degree$C)  & Reward & Average temperature ($\degree$C) & Humidity (\%) & Mean weather temperature deviation ($\degree$C) \\
		\midrule		
		Toronto	& 269.3	& 1.088	& -0.271 & 7.38	& 73.83	& 22.6\\
		Golden, CO & 224.9	& 1.287	& -0.1544 & 9.7	& 53.67	& 26\\
		Chicago, IL	& 350.1 & 1.48 & -0.1929	& 9.9 & 70.33 & 20.99\\
		Dulles, VA	& 393.5 & 1.231 & -0.1901 & 12.6 & 69.08 & 24.8\\
		San Francisco, CA & 227.3 & 1.032 & -0.1128	& 13.8 & 74.5 & 6.5\\
		Tampa, FL & 777.1 & 0.742 & -0.233 & 22.3 & 75.33 & 10.35\\
		\bottomrule
	\end{tabular}
\end{table}

\subsection{Comparison of Single vs Multi-agent}

Table~\ref{tab:result-multi-agent} provides the data used for generating Figure~\ref{fig:result-multi-agent}.

\begin{table}[H]
	\caption{Table comparing multi-zone vs. multi-agent policies on reward value, energy consumption, mean temperature violation, as $\alpha$ is varied.}
	\label{tab:result-multi-agent}
	\begin{tabular}{ccp{0.6in}p{0.6in}c}
		\toprule
		Control type & $alpha$ &  Energy consumption (MWh) & Mean temperature violation ($\degree$C & Reward)\\
		\midrule
		\multirow{9}{*}{Multi-zone} & 0.001	& 258.4 & 0.943 & -0.0709\\
		& 0.01	& 258.8 & 0.94	& -0.0721\\
		& 0.1 & 259.1 & 0.998	& -0.0861\\
		& 0.5 & 268.5 & 0.916	& -0.121\\
		& 1	& 258.6 & 0.846 & -0.2943\\
		& 5	& 263.2 & 0.783 & -0.767\\
		& 10 & 276.5 & 0.3571	& -0.6893\\
		& 100 & 301.9	& 0.028	& -0.3318\\
		& 1000 & 279.2	& 0.0261 & -2.875\\	
		\midrule								
		\multirow{9}{*}{Multi-agent} & 0.001 & 248.1 & 0.903 & -0.0702\\
		& 0.01 & 253.6	& 0.889	& -0.0712\\
		& 0.1 & 258.1 & 0.8642 & -0.0784\\
		& 0.5 & 257.1 & 0.8413 & -0.1425\\
		& 1	& 268.8 & 0.446 & -0.2355\\
		& 5	& 258.7 & 0.4652 & -0.4036\\
		& 10 & 263.2 & 0.51 & -0.6612\\
		& 100 & 275.2 & 0.022 & -0.2841\\
		& 1000 & 277.5 & 0.018 & -1.7213\\
		\bottomrule
	\end{tabular}
\end{table}

\end{document}